\documentclass[runningheads]{llncs}

\usepackage[utf8]{inputenc} 
\usepackage[T1]{fontenc}    
\usepackage{hyperref}       
\usepackage{url}            
\usepackage{booktabs}       
\usepackage{amsfonts}       
\usepackage{nicefrac}       
\usepackage{microtype}      
\usepackage{lipsum}
\usepackage{amsmath} 
\usepackage{amssymb} 
\usepackage{bm}
\usepackage{bbm}
\usepackage{graphicx}

\DeclareMathOperator{\E}{\mathbb{E}}

\usepackage{setspace}

\begin{document}
\title{Sampling Techniques in Bayesian Target Encoding}
\titlerunning{Sampling Bayesian Encoder}
%
%
\author{Michael~Larionov}
%
%
\institute{Spok,~Inc. \email{michael.larionov@spok.com}}

\maketitle

\begin{abstract}
Target encoding is an effective encoding technique of categorical variables and is often used in machine learning systems for processing tabular data sets with mixed numeric and categorical variables. Recently en enhanced version of this encoding technique was proposed by using conjugate Bayesian modeling. This paper presents a further development of Bayesian encoding method by using sampling techniques, which helps in extracting information from intra-category distribution of the target variable, improves generalization and reduces target leakage.
\end{abstract}

\section{Introduction}

Target encoding technique was formulated first in \cite{targetencodingpaper} as a way to deal with high-cardinality categorical variables. In this technique each value of the categorical variable (which we call for simplicity a category) is mapped to a target mean conditional on the value of the variable. More precisely, for regression problem it is an expected value of the target given the category. For binary classification it is posterior probability of the target given the category. For multi-class problem an obvious extension of the binary case would be to introduce m-1 new variables (where m is the number of classes) that are the posterior probabilities of the target being in a specified class. 

This technique proved remarkably successful in a variety of machine learning projects and became extremely popular in data science competitions, such as those offered on the Kaggle platform\footnote{https://www.kaggle.com}. It provides better model performance than 1-hot encoding because it uses target statistics to create encoding variables, thus making it more meaningful than 1-hot encoding that treats all values of the categorical variable equally\cite{survey2020}. 
For several different implementations of Target Encoding see python package Category Encoders\footnote{http://contrib.scikit-learn.org/category\_encoders} \cite{catencoders}. Target Encoding is also widely known as Mean Encoding because it encodes categorical variables with conditional target mean. This and some other methods of categorical encoding are compared in the study \cite{survey2017}.

One of the issues with Target Encoding is that it fails to extract information from intra-category target variable distribution because it is using only the target mean, leaving out other relevant statistics. This shortcoming is addressed in Bayesian Target Encoding techniques \cite{CBM,betaencoder}, the main idea of which is to select a conjugate prior for the conditional distribution of the target variable given the value of the categorical variable, and then update the parameters of the conjugate prior based on training examples to obtain a posterior distribution, and encode the categorical variable using the first several moments of the posterior distribution. Slakey et al. \cite{CBM} have the encoding of the categorical variable done in two steps or layers: the Local Layer in which the posterior distribution is computed for each value of the categorical variable, and the Encoding Layer where each category is encoded using the first Q moments of the posterior distribution.

In this paper we propose to enhance the Encoding Layer by sampling from the posterior distribution instead of taking expectations of its first Q moments of its parameters. We call this method Sampling Bayesian Encoder. It puts the Bayesian Target encoding on a solid theoretical ground, opening window for more improvement of the target encoding techniques. It eliminates the need to add Gaussian noise to the encoded values, as is commonly done to avoid target leakage and overfitting \cite{catencoders}, because we are exploring the parameter space during sampling procedures.  This approach was used on simulated data as well as on real world data and demonstrated better generalization for the data sets with a mixture of numeric and categorical variables with various degree of cardinality.

\section{Sampling as embedding layer}

Following Slakey et al. \cite{CBM}, we train a probabilistic model of the target variable $y$ for each categorical variable, deriving a posterior distribution:

\begin{equation} \label{eq:posterior}
    p_{mv}(\boldsymbol\theta|y) \propto \mathcal{L}_{mv}(\boldsymbol\theta|y)p(\boldsymbol\theta),
\end{equation}

where $v$ signifies unique values of the categorical variable $x_m$, $\boldsymbol\theta$ is a parameter vector of the posterior distribution, $p_{m}(\boldsymbol\theta)$ is a prior distribution, and $\mathcal{L}_{mv}(\boldsymbol\theta|y)$ is a likelihood function.

Let us define $\boldsymbol\theta_{nm}$ as a realization of the posterior distribution $p_{mv}(\boldsymbol\theta|y)$ for categorical variable m and example n:

\begin{equation}
    \boldsymbol\theta_{nm} \sim p_{mv}(\boldsymbol\theta|y) \text{, where   } v = x_{nm}
\end{equation}

Let us also introduce a vector function $\boldsymbol f_m(\boldsymbol\theta)$ that maps the parameters of the posterior distribution to a Q-dimensional space and encode all categorical variables $x_m$ with $\boldsymbol f_m$ and train a model $\hat{y}_{\theta}(\xi_n, \boldsymbol f_1(\boldsymbol\theta_{n1})..\boldsymbol f_M(\boldsymbol\theta_{nM}))$ using encoded categorical variables and numeric variables $\xi_n$ that do not require encoding. We do not place any restriction on the algorithm to train the model $\hat{y}_{\theta}$. It could be a linear model, Random Forest, Gradient Boosted Trees, SVM, a Neural network, or any other algorithm.

Since $\boldsymbol\theta_{nm}$ is a random variable, we will train the model $\hat{y}_{\theta}$ for all possible values of $\boldsymbol\theta_{nm}$ and take an expectation of the target variable to get an expected prediction:

\begin{equation}\label{eq:expectation}
    \hat{y}(x_n) = \E_{\boldsymbol\theta_{nm} \sim p_{mv}(\boldsymbol\theta|y)}\hat{y}_{\theta}(\xi_n, \boldsymbol f_1(\boldsymbol\theta_{n1})..\boldsymbol f_M(\boldsymbol\theta_{nM}))
\end{equation}

The choice of mapping function $\boldsymbol f_m(\boldsymbol\theta)$ is more important for less expressive models and less so for more expressive models that can learn this function during training process. Examples of this function include identity function (like in most Target Encoding schemes), polynomials of $\boldsymbol\theta$ (more in line with \cite{CBM,betaencoder}), or more complex functions, for example, weight of evidence.

For all but very simple models the expectation (\ref{eq:expectation}) is intractable, but it can be approximated by sampling from the posterior distribution. For a sample $\boldsymbol\theta_{nm}^k$ where $k=1..K$, the estimate is:

\begin{equation} \label{eq:estimate}
    \bar{y}(x_n) \approx \frac{1}{K}\sum_{k=1}^K{\hat{y}_{\theta}(\xi_n, f_1(\boldsymbol\theta_{n1}^k)..f_M(\boldsymbol\theta_{nM}^k))}
\end{equation}

Sampling from the posterior distribution can be done at each epoch provided the algorithm to predict $\hat{y}_{\theta}$ supports it. Otherwise after sampling $K$ times we encode the categorical variables with the realizations $\boldsymbol\theta_{nm}^k$ and concatenate the encoded data together. Thus, if the original training set contains $N$ examples, then after encoding we will get $KN$ examples on which we train the model $\hat{y}_{\theta}$. During prediction phase we will also  sample from the posterior distribution and then average the predictions of the model $\hat{y}_{\theta}$ and for previously unseen categories we use values sampled from the prior distribution.

Following \cite{CBM}, we compute the prior distribution using target statistics of the entire data set. However, unlike \cite{CBM}, we scale down the parameters of the prior distribution as if they were computed on a subset of the training data using a scaling factor, which is a hyperparameter of the model. Scaling down the influence of the prior distribution helps control better bias-variance trade-off for rare categories.

\section{Practical implementation of the sampling techniques}

Model training for regression, binary classification and multiclass classification problems will follow the same steps:
\begin{enumerate}
    \item \label{step:prior} Finding a prior distribution $p(\boldsymbol\theta)$ by using target statistics for the entire training data, then scaling it down to reduce its excessive influence on the results.
    \item \label{step:posterior} Finding a conditional posterior distribution $p_{mv}(\boldsymbol\theta|y)$ for each categorical variable and for each value of the categorical variable. If we are using conjugate priors the posterior distributions can be found analytically. This step is identical to the Local Layer in \cite{CBM}
    \item Generating an augmented set that contains $K$ copies of the training set with all categorical features encoded using samples from the posterior distributions
    \item Training a model (Random Forest, SVM, etc.) on the augmented set
    \item Finding prediction by averaging of $K$ results of the models with $K$ different encoded values $\boldsymbol\theta$.
\end{enumerate}

\subsection{Binary classification tasks}

Binary classification is the simplest case because of the obvious choice of the target variable distribution: Bernoulli distribution. The conjugate prior for Bernoulli distribution is Beta distribution that has two parameters $\alpha$ and $\beta$. In Beta-Binomial model these parameters have a simple interpretation as $\alpha-1$ successes and $\beta-1$ failures. So during step \ref{step:prior} we set the parameters of the prior distribution as follows:

\begin{equation}
    \alpha = 1 + \gamma \sum_{n=1}^N y_n
\end{equation}

\begin{equation}
    \beta = 1 + \gamma \sum_{n=1}^N (1-y_n),
\end{equation}

where $\gamma$ is a non-negative scaling factor and is a hyperparameter to the model. Zero value of $\gamma$ indicates an uninformative prior that does not use any target statistics, and  the greater the value is, the more the marginal target statistics influence the encoding of the categorical features. 

During step \ref{step:posterior} the parameters of the posterior distribution is updated for every category:

\begin{equation}
    \alpha_{mv} = \alpha + \sum_{x_nm=v} y_n
\end{equation}

\begin{equation}
    \beta_{mv} = \beta + \sum_{x_nm=v} (1-y_n)
\end{equation}

More frequent categories will give us higher values of $\alpha$ and $\beta$ and result in a sharper peak in the Beta distribution, so most of the samples will be around its maximum value. For the infrequent categories the values of $\alpha$ and $\beta$ will be lower and the distribution will be wider, resulting in greater variance of the samples. For very infrequent categories the distribution will be very close to the prior. This prevents the model from overfitting on potentially extreme target values for rare categories. There is also no need to add Gaussian noise to the encoded values, because by virtue of the sampling technique the target leakage is greatly reduced.

\begin{figure*}[!htb]
    \center{
        \vspace*{0.5in}
        \includegraphics[width=0.4\textwidth]{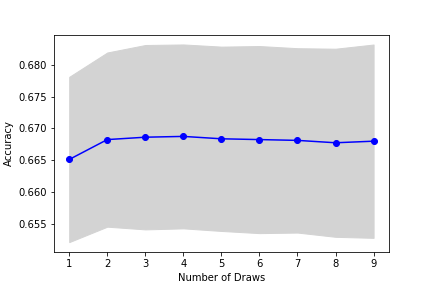} 
        \includegraphics[width=0.4\textwidth]{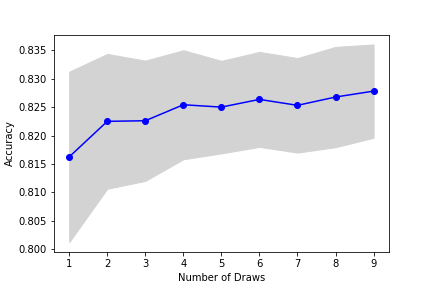} 
    }
    \caption{\label{fig:draws} Model performance for make\_classification() and  make\_hastie\_10\_2() data depending on number of draws. Shaded area indicates 1$\sigma$ prediction intervals}
\end{figure*}

\begin{figure*}[!htb]
    \center{
        \includegraphics[width=0.4\textwidth]{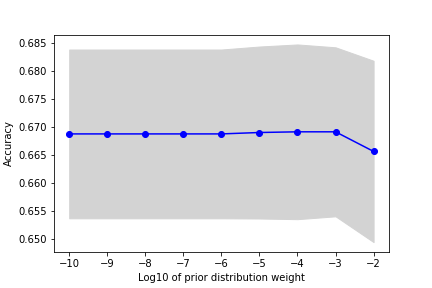} 
        \includegraphics[width=0.4\textwidth]{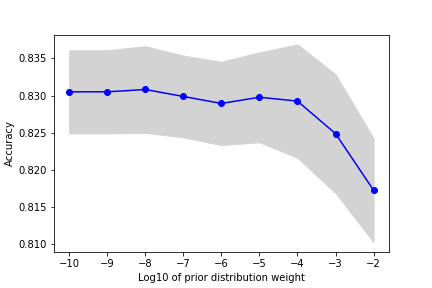} 
    }
    \caption{\label{fig:prior} Model performance for make\_classification() and  make\_hastie\_10\_2() data depending on $log_10$ of prior distribution rate. Shaded area indicates 1$\sigma$ prediction intervals
}
\end{figure*}

\subsection{Multi-class classification tasks}
Multi-class classification task is an extension of binary classification, and so are the probability distributions used for the category encoding task. The target variable is described by Categorical distribution, and the conjugate prior is Dirichlet distribution, the parameter of which is a vector $\boldsymbol{\alpha}$ of the same dimension as the number of classes, where all components are greater than zero. We set the prior as:

\begin{equation}
    \alpha^c = 1 + \gamma \sum_{n=1}^N \mathbbm{1}(y_n=c)
\end{equation}

and the parameters of the posterior distribution will be updated as follows:

\begin{equation}
    \alpha_{mv}^c = \alpha^c +  \sum_{x_{nm=v}} \mathbbm{1}(y_n=c)
\end{equation}

\subsection{Regression tasks}\label{section:regression}

Regression case is more complicated, because the continuous target variable can rarely be modeled using the same type of distribution for all categorical variables. But in the simplest case it can be modeled as a Normal distribution. This distribution has two parameters: $\mu$ and $\sigma^2$, or alternatively $\mu$ and precision $\tau=\sigma^{-2}$. The prior distribution is Normal-Inverse Gamma and Normal Gamma respectively. Both of them have the same set of parameters $\mu_0$, $\nu$, $\alpha$, $\beta$. The parameters of the prior distribution is:

\begin{equation}
    \mu_0 =\bar{y} = \frac{1}{N}\sum_{n=1}^N y_n
\end{equation}

\begin{equation}
    \nu = 0
\end{equation}

\begin{equation}
    \alpha = \gamma \frac{N}{2}
\end{equation}

\begin{equation}
    \beta = \frac{\gamma}{2} \sum_{n=1}^N (y_n - \bar{y})^2
\end{equation}

The posterior distribution for each category is:

\begin{equation}
    \mu_{0mv} = \frac{\nu\mu_0 + N_{mv}\bar{y}_{mv}}{\nu+N_{mv}}
\end{equation}
\begin{equation}
    \nu_{mv} = \nu + N_{mv}
\end{equation}
\begin{equation}
    \alpha_{mv} = \alpha + \frac{N_{mv}}{2}
\end{equation}
\begin{equation}
\begin{split}
    \beta_{mv} = \beta + \frac{1}{2}\sum_{n=1}^{N_{mv}}{(y_n - \bar{y}_{mv})^2} +  \frac{N_{mv}\nu}{\nu+N_{mv}}\frac{(\bar{y}_{mv} - \mu_0)^2}{2}, 
\end{split}
\end{equation}

where $\bar{y}_{mv}$ is the sample mean of the target values for the category mv. 

While it is possible to come up with more complex conditional target distributions, for example, Gaussian Mixture models, Normal is usually works pretty well and an improvement against the deterministic Target Encoding techniques.

\begin{figure*}[!htb]
    \center{
        \vspace*{0.5in}
    \includegraphics[width=\textwidth]{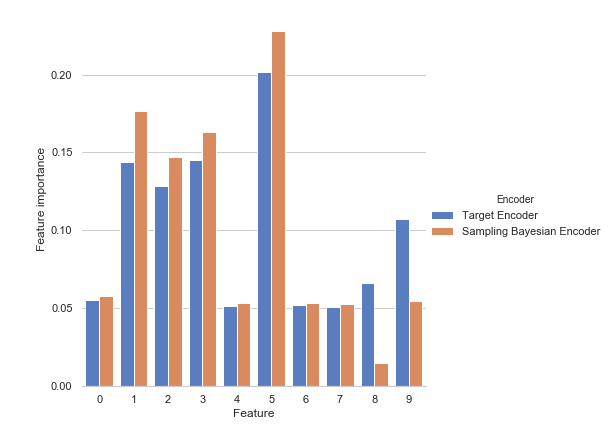} 
    }
    \caption{\label{fig:import} Feature importance for make\_classification() data for the models using Target encoder and Sampling Bayesian encoder as reported by Random Forest algorithm. Features 0-7 are numeric, and features 8 and 9 are categorical. Sampling Bayesian encoder makes the training algorithm put less emphasis on the categorical features thus reducing target leakage}
\end{figure*}

\section{Related work}

The problem of target leakage was discussed in details in \cite{catboost}, where a new  sampling technique  called Ordered Target Statistics was proposed. The training data are reshuffled and for each example the categorical features are encoded with the target statistics of all previous entries. Thus the "earlier" examples have a higher variance than the "later" examples, and to deal with this issue several permutations are taken and one permuted set is picked at random for every iteration of the Gradient Boosted Trees algorithm.  The idea of taking multiple permutations is similar to our idea of generating multiple samples from the posterior distribution, however the distribution from which the sampling occurs in \cite{catboost} is not the posterior distribution, but the sampling distribution averaged over all possible sample sizes. It can be shown that this distribution has the same mean that the posterior distribution, but an elevated variance, and this can be detrimental to model expressiveness.

LightGBM categorical encoding also uses target statistics, but it deals with target leakage by finding the best split based on Fisher's method of minimizing intra-category variance \cite{fisher,LightGBM1,LightGBM2}. See Appendix \ref{appendix:a} for a explanation how the reduction of intra-category variance results naturally from the sampling technique proposed in this paper.

\section{Experimental Studies}

We designed our experiments with two goals in mind: to compare model performance for different kinds of category encoding and, specifically for Sampling Bayesian encoder, learn how hyperparameters affect the model performance. For this reason we used both synthetic and real world data.

\subsection{Data}

We generated synthetic data sets using  \verb|skikit-learn| functions \verb|make_classification()| and \verb|make_hastie_10_2()|. In both cases the data set contains 10 columns and 10,000 rows and several columns were converted to categorical variables using \verb|KBinsDiscretizer| with the number of categories ranging from 10 to 20. 
We left the other variables numeric to compare the relative feature importance of the categorical and numeric variables encoded using deterministic vs. sampling approach.

For a regression problem we used a past Kaggle competition "Mercedes-Benz Greener Manufacturing" \footnote{https://www.kaggle.com/c/mercedes-benz-greener-manufacturing/}. This data set has 4209 rows and 385 independent variables, ten of which are high-cardinality categorical variables, for which, as we know from \cite{targetencodingpaper,CBM}, 1-hot encoding does not produce good results. This makes it a good candidate for Target Encoding and any versions of it, including our own. The learning task is to predict the numeric variable y and the evaluation metric is $R^2$

Another real world data set is Adult data\footnote{https://archive.ics.uci.edu/ml/machine-learning-databases/adult/adult.data} that was extracted from the 1994 Census Bureau\cite{CBM}. This data set contains six numeric  and eight categorical variables and  32561 rows. The learning task is to predict whether or not someone earns more than \$50,000. We used accuracy as an evaluation metric.

\subsection{Hyperparameters}

Hyperparameter tuning was done using  5-fold cross-validation for both baseline and sampling encoding models. After obtaining the best model from the cross-validation procedure we varied one hyperparameter of the encoder at a time to see how it influences the model performance. Random Forest estimators also provided feature importances of the trained models. We tuned hyperparameters for both our technique and the techniques with which we compare the model performance.

Sampling approach was compared to the Target Encoding as implemented in \verb|LeaveOneOutEncoder| class of \verb|category_encoders| package \cite{catencoders} and CBM encoder as reported in \cite{CBM} for the Adult data set. To select the mapping function $f()$ we tried several options. For regression tasks these were using mean only (as in \cite{targetencodingpaper}), using both mean and precision and adding second-degree polynomial of these variables. For binary classification tasks we also added weight of evidence to the list of possible mapping functions. The more expressive is the algorithm the less necessary is to pick the mapping function $f()$ as the algorithm is able to approximate it during training. This is confirmed on the two simulation studies, in which Random Forest was used. For less capable algorithm, for example, logistic regression, manual feature engineering becomes more important and the correct pick of the mapping function could dramatically improve model performance.

Another important hyperparameter is the prior scaling factor $\gamma$, which is varied from zero (meaning uninformative prior) to one (meaning that the prior distribution has the same degree of influence on the posterior distribution as the likelihood function. Experimental studies on synthetic data (Figure \ref{fig:prior}) show that this parameter exerts little influence on the model performance unless it is too big, at which point performance of the model deteriorates. This is probably the case because the prior distribution is effective on correcting overfitting for infrequent categories, but these categories contribute little to the overall loss function. 

Finally, we can also select the number of samples $K$ in \ref{eq:estimate}. Small values of $K$ may result in reduced accuracy, while large values of $K$ make the training process slow. Experimental studies on the synthetic data sets demonstrates that large samples are not required for good model performance. For \verb|make_classification()| data (Figure \ref{fig:draws}, left) the  accuracy is unchanged after 2. For \verb|make_hastie_10_2| (Figure \ref{fig:draws}, right) the accuracy slowly increases when $K$ is greater than 2. This can be explained by the fact that all categories have large number of observations, so all the samples from the posterior distribution are close to its mode. In this situation large sample approximation (Appendix \ref{appendix:b}) may be applied. The real world data often have infrequent categories, for which the posterior distribution is wide, so a higher number $K$ may be required to achieve optimal model performance. 

\subsection{Results and Discussion}    

Models using Sampling Bayesian Encoder consistently show better results than those using other encoding techniques. For example, the best model using the proposed technique and Random Forest classifier achieved accuracy of 0.6595 on a \verb|make_classification()| data with one numeric and nine categorical features vs. 0.6585 for the best model using \verb|LeaveOneOutEncoder|. Same experiment on \verb|make_hastie_10_2| produced 0.8229 and 0.8217 respectively. This edge is preserved on the real world data. For example, on Adult data set the new approach produced accuracy of 0.8652 vs the best model accuracy of 0.8647  for the models using \verb|LeaveOneOutEncoder| (Slakey et al. \cite{CBM} reported 0.86 for \verb|RandomForestClassifier|). On Mercedes-Benz data the $R^2$ of the models using our approach is 0.61135 vs. 0.60974 for \verb|LeaveOneOutEncoder|. Moreover, when we submitted the results to the competition we've got improved scores on both public and private data sets.

This demonstrates that the models using Sampling Bayesian Encoder can generalize well on the unseen data. Despite improved accuracy, we observe slowness in the training procedure, primarily because we need to sample from Beta, Dirichlet or Gamma distributions, and this adds to resource consumption during the data preprocessing stage, but also because it effectively increases the size of the data set depending on the parameter $K$, which also slows down training procedure. 

Many traditional encoders of the categorical variables are causing the models to put too much importance on the categorical features, which result in poor generalization. To see how Sampling Bayesian Encoder handles this problem we compared feature importances of \verb|RandomForestClassifier| for the models with Target Encoder and Sampling Bayesian encoder. The results are shown in Figure \ref{fig:import}. It is indeed clear that Sampling Bayesian Encoder cause the classifier to put much less importance on the categorical variables (labeled 8 and 9), thus reducing target leakage.

The source code of the experiments is available at the project's \verb|github| repository\footnote{ https://github.com/mlarionov/sampling\_bayesian\_encoder}.

\section{Conclusion}

In this paper we presented a new technique of categorical variable encoding by sampling from the conditional posterior distribution of the target variable given the value of the categorical variable.
This method is a logical development  of the target encoding methods represented in \cite{targetencodingpaper}, \cite{CBM}, and \cite{betaencoder} and is capable of  reducing the propensity of Target Encoding and Bayesian Target Encoding to overfit due to the target leakage into the predictor data. 

\bibliographystyle{unsrt}  
\bibliography{references}

\appendix
\section{Appendices}

\subsection{Regularization effect of sampling techniques}\label{appendix:a}

In this section we consider how the loss function is modified by virtue of using sampling techniques in target encoding of the categorical variable.  We will consider a regression case already discussed in section \ref{section:regression} and Mean Squared Error (MSE) loss function, which is used to optimize model $\hat{y}_\theta$:

\begin{equation}\label{eq:mse}
\begin{split}
MSE = \frac{1}{N} \sum_n \E_{\boldsymbol\theta_{nm} \sim p_{mv}(\boldsymbol\theta|y)}\big(y_n-
\hat{y}_{\theta}(\xi_n, \boldsymbol f_1(\boldsymbol\theta_{n1}) ..\boldsymbol f_M(\boldsymbol\theta_{nM}))\big)^2    
\end{split}
\end{equation}

Expanding the square in (\ref{eq:mse}) and taking expectation of each term, we can represent MSE as a sum of two terms:

\begin{equation}
    MSE = MSE_0 + REG,
\end{equation}

where $MSE_0$ is the Mean Squared Error of the point estimate:

\begin{equation}
    MSE_0 = \frac{1}{N} \sum_n \big(y_n-\hat{y}(x_n)\big)^2    
\end{equation}

and $REG$ is a regularization term:

\begin{equation}
\begin{split}
    REG = \frac{1}{N} \sum_n \E_{\boldsymbol \theta_{nm} \sim p_{mv}(\boldsymbol\theta|y)} 
    \big(\hat{y}_{\theta}(\xi_n, \boldsymbol f_1(\boldsymbol\theta_{n1}) ..\boldsymbol f_M(\boldsymbol\theta_{nM}))-\hat{y}(x_n)\big)^2    
\end{split}
\end{equation}

It represents average difference between $\hat{y}_f$ and the average $\hat{y}_q$ under the encoding distribution $g$. This means that our optimization objective favors models that produce small variance of predictions within the categories. This is in line with LightGBM approach of minimizing intra-category variance, but we have it here as a natural consequence of the sampling technique.

\subsection{Large sample approximation}\label{appendix:b}

We can use the property of the large samples, that the parameters of the posterior distribution follow approximately Normal distribution around its maximum \textit{a posteriori} estimation \cite{LittleRubin}:

\begin{equation}
    (\boldsymbol\theta - \hat{\boldsymbol\theta}) \sim N(0, \boldsymbol C)
\end{equation}

The covariance matrix $C$ is derived by Taylor series expansion of the logarithm of the posterior distribution and is the inverse of the information matrix. Applying Taylor series expansion to $\hat{y}_\theta$:

\begin{equation}\label{f_taylor}
    \hat{y}_\theta(\xi, \boldsymbol\theta) \approx \hat{y}_\theta(\xi, \hat{\boldsymbol\theta}) + \nabla_\theta\hat{y}_\theta \delta \boldsymbol\theta + \frac{1}{2} \delta \boldsymbol\theta ^\intercal \mathbf{H}(\hat{y}_\theta) \delta \boldsymbol\theta,
\end{equation}

where $\hat{\boldsymbol\theta} = \E_{\theta \sim p(\theta|y)}\boldsymbol\theta$ is an expected parameter of the posterior distribution,  $\delta\boldsymbol\theta = \boldsymbol\theta - \hat{\boldsymbol\theta}$ and $\mathbf{H}(\hat{y}_\theta)$ is a Hessian matrix. Taking expectation of (\ref{f_taylor}) with respect to the posterior, we get an estimate of $\hat{y}$:

\begin{equation}
    \hat{y}(x) \approx \hat{y}_\theta(\xi, \hat{\boldsymbol\theta}) + \frac{1}{2} \mathbf{tr} (\mathbf{H}(\hat{y}_\theta) ~\boldsymbol C)
\end{equation}

The first term can be interpreted as a Target encoding estimate, and the second term is an estimate correction due to intra-category variance.

We can contrast this with a popular technique of adding a Gaussian noise to the maximum \textit{a posteriori} estimate. Both approaches will produce the same results in large sample approximation if the intra-category variance is the same for all categories. While traditional Target Encoder requires a hyperparameter to control the Gaussian noise, the Sampling Target Encoder learns the variance from data, and can produce better results when this variance is different for different category values.

\end{document}